\documentclass{article}

\usepackage{kr}

\usepackage{times}
\usepackage{soul}
\usepackage{url}
\usepackage[hidelinks]{hyperref}
\usepackage[utf8]{inputenc}
\usepackage[small]{caption}
\usepackage{graphicx}
\usepackage{amsmath}
\usepackage{amsthm}
\usepackage{booktabs}
\usepackage{algorithm}
\usepackage{listings}
\usepackage[noend]{algpseudocode}
\usepackage{bussproofs}
\usepackage{mathrsfs}
\usepackage{mathtools}

\usepackage{makecell}

\usepackage{todonotes}

\graphicspath{{media/}}     

\title{Neuro-symbolic Commonsense Social Reasoning}

\author{%
David Chanin   
\and
Anthony Hunter\\
\affiliations
Department of Computer Science\\
University College London\\
\emails
\{david.chanin.22, anthony.hunter\}@ucl.ac.uk
}

\begin{document}
\maketitle

\begin{abstract}
Social norms underlie all human social interactions, yet formalizing and reasoning with them remains a major challenge for AI systems. We present a novel system for taking social rules of thumb (ROTs) in natural language from the Social Chemistry 101 dataset and converting them to first-order logic where reasoning is performed using a neuro-symbolic theorem prover. We accomplish this in several steps. First, ROTs are converted into Abstract Meaning Representation (AMR), which is a graphical representation of the concepts in a sentence, and align the AMR with RoBERTa embeddings. We then generate alternate simplified versions of the AMR via a novel algorithm, recombining and merging embeddings for added robustness against different wordings of text, and incorrect AMR parses. The AMR is then converted into first-order logic, and is queried with a neuro-symbolic theorem prover. The goal of this paper is to develop and evaluate a neuro-symbolic method which performs explicit reasoning about social situations in a logical form.
\end{abstract}

\section{Introduction}

Deep learning approaches have seen success on a wide range of natural language understanding (NLU) and natural language processing (NLP) tasks \cite{vaswani2017attention}. However, these models are largely black boxes and it is difficult to understand how these models arrive at conclusions. This is especially problematic when dealing with controversial social norms, where current deep learning models have been shown to learn a range of problematic racial and gender biases \cite{abid2021persistent,bolukbasi2016man,lu2020gender}.
It is not hard to envision how this bias will lead to social harm as we rely more and more on language models to generate content and power systems that make decisions that affect people's lives.

While interpretability methods exist that can be applied to deep learning models \cite{lime,NIPS2017_7062,kim2018interpretability}, these methods at best provide hints into what sorts of features influence a model's predictions, but cannot directly explain the reasoning process the model took internally, and do not provide a way to directly change problematic model behavior.

Symbolic logic approaches, on the other hand, reason with explicitly defined rules which gives a number of benefits for interpretability, multi-hop reasoning, and directly changing model behavior. It is possible to trace through inference steps in a symbolic logic system to determine how a conclusion was derived, and tweak any problematic logical formulae as necessary.  Ideally, we should be able to vet the logical rules that a model is using during inference and delete or modify any undesirable rules as needed.

Symbolic logic, however, tends to be brittle and struggles with small inconsistencies in input data, as symbols in logical formulae cannot represent semantically similar concepts like ``dad" and ``father" without every possible alternate symbol and relation being explicitly added to the system. This brittleness is not a problem for deep learning, however, as the entire system learns to simply maximize output over noisy inputs. These contrasting strengths motivate neuro-symbolic approaches which can combine the strengths of both symbolic logic and deep learning.

The goal of this work is to develop a neuro-symbolic social reasoning system to do logical reasoning using social rules of thumb described in natural language. We accomplish this in several steps.
First, we leverage Abstract Meaning Representation (AMR) \cite{banarescu2013abstract} by converting
natural language social rules-of-thumb into AMR and then align them with RoBERTa embeddings. We then generate alternate simplified versions of the AMR for robustness against different wordings of ``semantically equivalent" sentences and AMR parse errors. The AMR is then converted into first-order logic, and is queried with a neuro-symbolic theorem prover \cite{rocktaschel2017end,kalyanpur2022braid}.

We use natural language social rules-of-thumb from the Social Chemistry 101 dataset \cite{forbes2020social} as a source of rules for the reasoning system. Social Chemistry 101 is a dataset containing over 290,000 rules of thumb (ROTs) for evaluating people's behavior in everyday social situations. Social Chemistry ROTs are drawn from Reddit's \texttt{r/confessions} and \texttt{r/amitheasshole} subreddits, the ROCStories corpus \cite{mostafazadeh2016corpus}, and from the Dear Abby advice column web archives \footnote{https://www.uexpress.com/life/dearabby/archives}. These rules of thumb cover a wide range of common situations, with ROTs like ``You should be there for your child's birthday" or ``It is wrong to damage someone else's property on purpose". ROTs consist of a short sentence containing a judgement like ``It's wrong to ..." or ``It's reasonable to ..." followed by a description of the situation being judged. In addition, the dataset contains a social situation text (SST) for each ROT which gives an example that the ROT applies to. A sample of Social Chemistry 101 data is shown in Table \ref{tab:selectedrots}.

These ROTs have the potential to be a great source of commonsense knowledge for a reasoning system, if they can be parsed into a logical form that makes it possible to do reasoning. This paper attempts to do exactly this, by turning Social Chemistry ROTs into first-order logic, as will be described in more detail in section \ref{sec:approach}.

\begin{table}
    \begin{center}
        \vskip5mm
        {\renewcommand{\arraystretch}{1.2}%
        \begin{tabular}{| p{3.5cm} | p{3.5cm} |} 
            \hline
            \thead{Rule of Thumb} & \thead{Social Situation Text} \\ [0.5ex] 
            \hline
            You shouldn't put dog poop in front of someone's door & putting my neighbor's dog's poop in front of their door \\ 
            \hline
            It is wrong to be jealous of your partner & being jealous of my gf \\
            \hline
            It's expected that you'll buy Christmas presents for the kids in your life & not buying my roommate's kids Christmas presents this year \\
            \hline
            It's bad to be an enabler of harmful behavior & not wanting to enable my grandparents hurting themselves \\
            \hline
        \end{tabular}} \quad
        \vskip3mm
        \caption{Sample Social Chemistry ROTs and corresponding SSTs}
        \label{tab:selectedrots}
    \end{center}
\end{table}

We evaluate the performance of our method on Social Chemistry 101 by measuring the degree to which the method is able to match each ROT in the dataset to its corresponding SST via logical inference, and verify as well that we should not be able to conclude a match via logical inference when applying each ROT to a different, randomly chosen SST from the dataset. For example, we would expect the ROT ``It is wrong to be jealous of your partner" to apply to the SST ``being jealous of my girlfriend", but it should not apply to the SST ``Not buying my roommates' kids Christmas presents".

We proceed as follows: In Section \ref{sec:amr}, we give a overview on AMR and converting AMR into logical formulae; In Section \ref{sec:theoremproving}, we give a background overview on neuro-symbolic theorem proving; In Section \ref{sec:approach}, we detail our social reasoning system; In Section \ref{sec:eval} we evaluate our technique against the Social Chemistry dataset; In Section \ref{sec:relatedwork}, we discuss related work; In Section \ref{sec:discussion} we discuss future directions for this work. The code for this paper is available online \footnote{\href{https://github.com/chanind/amr-social-chemistry-reasoner}{https://github.com/chanind/amr-social-chemistry-reasoner}}.

\section{Abstract meaning representation}
\label{sec:amr}

Abstract meaning representation (AMR) \cite{banarescu2013abstract} is a powerful abstraction which can help simplify and standardize sentences for semantic meaning. AMR represents text as a rooted, directed acyclic graph drawing predicate senses and semantic roles from the OntoNotes project \cite{hovy2006ontonotes}. An example is shown in Figure \ref{fig:basicamr}.

\begin{figure}
    \footnotesize
    \begin{verbatim}
    (w / want-01
        :ARG0 (b / boy)
        :ARG1 (b2 / believe-01
            :ARG0 (g / girl)
            :ARG1 b))
    \end{verbatim}
    \vspace{-5mm}
    \caption{AMR for the sentence ``The boy wants the girl to believe him"}
    \label{fig:basicamr}
\end{figure}

AMR has first-class support for negation, which is particularly valuable when translating AMR into first-order logic. This is demonstrated below for the sentence ``The boy does not go", where negation is represented via the :polarity relation.

\begin{footnotesize}
\begin{verbatim}
(g / go-02
    :ARG0 (b / boy)
    :polarity -)
\end{verbatim}
\end{footnotesize}

The numbers after the instance name in AMR (e.g. go-02 above) refers to a specific OntoNotes or PropBank \cite{kingsbury2002treebank} semantic frame. These frames can take different parameters, but in general \texttt{:ARG0} refers to the subject and \texttt{:ARG1} refers to the object. In our work, however, we find that these frame numbers (the -02 in go-02) tend to get mixed-up by existing AMR parsers, so we discard these frame numbers when they appear in AMR. For the rest of this paper, we will remove frame numbers in AMR.

AMR makes a number of simplifying assumptions in order to represent semantically similar sentences using identical structure. AMR does not differentiate between verbs and nouns, and cannot represent verb tenses. It does not represent singular and plural, quotations, or articles. For instance, the AMR in Figure \ref{fig:basicamr} could refer to any of the following English sentences, among many others: ``The boy desires the girl to believe him.", ``The boy will desire the girl to believe him.", ``A boy desires a girl to believe him.", ``The boy has a desire to be believed by the girl.", etc... For this research, the simplification of English that AMR performs is generally beneficial since it removes information from English sentences which is less relevant to the task at hand.

AMR was also chosen due to the availability of high-quality off-the-shelf AMR parsers \cite{lee2021maximum,drozdov2022inducing} and tooling like the Python Penman library \cite{goodman-2020-penman}.

\begin{figure}[ht]
\centerline{\includegraphics[width=\linewidth]{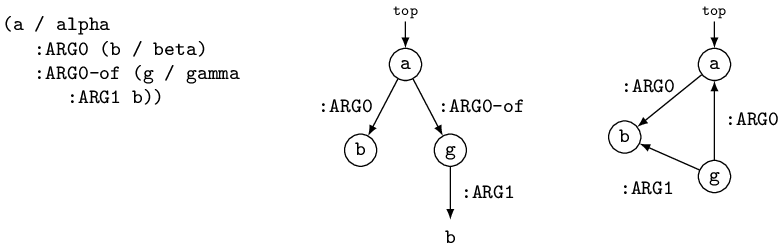}}
\caption{AMR represented as text, a tree, and a graph, from the Python Penman library.}
\label{fig:amrtreegraph}
\end{figure}

AMR can be represented as either a tree or a graph, with the tree directly mapping to the AMR text format \cite{goodman-2020-penman}. An example of showing the same AMR represented as text, a tree, and a graph is shown in \ref{fig:amrtreegraph}. It is always possible to generate the graph representation from the tree representation, but the reverse is not always possible as there are multiple possible trees that map to the same graph. In the rest of this paper, we always use the tree representation rather than the graph representation when working with AMR.

AMR also benefits from existing work on converting AMR to first-order logic. \cite{bos2016expressive} describes an algorithm for converting an AMR graph into existentially-quantified first-order logic joining all predicates by an implicit conjunction. An example of this is shown below, where the AMR for ``The boy does not go" from Figure \ref{fig:negamr} is converted into logical formulae as follows:
   \[
   \neg \exists G.(\textrm{go}(G) \land \exists B.(\textrm{:ARG0}(G, B) \land \textrm{boy}(B)))
   \]

Another example of AMR and its converted logical formulae from \cite{bos2016expressive} is show in Figure \ref{fig:amrlogbos}.


\begin{figure}[ht]
    \footnotesize
    \begin{verbatim}
(e / dry
    :ARG0 (x / person
        :named "Mr Krupp")
    :ARG1 x)
    \end{verbatim}
    \vspace{-5mm}
    \begin{align}
        & \exists X.(\textrm{person}(X) \land \textrm{named}(X, \textrm{``Mr Krupp"}) \nonumber\\
        & \hspace{5mm} \land \; \exists E.(\textrm{dry}(X) \land \textrm{:ARG0}(E,X) \land \textrm{:ARG1}(E,X) )) \nonumber
    \end{align}
    \caption{AMR and corresponding logical formulae representation of the AMR for ``Mr Krupp dries himself"}
    \label{fig:amrlogbos}
\end{figure}

There is not space to reproduce the full algorithm of \cite{bos2016expressive} here, but we refer the reader to the original paper for details. In addition, we implemented this algorithm in an open-source Python library called ``Amr Logic Converter" \footnote{\href{https://github.com/chanind/amr-logic-converter}{https://github.com/chanind/amr-logic-converter}} which is used in this paper.

\section{Resolution with non-binary unification}
\label{sec:theoremproving}

Traditional theorem provers unify predicates using an exact match on the name of the predicate \cite{ertel2018introduction}. For instance, the statement \texttt{father(Homer, Bart)} would unify against \texttt{father(Homer, Y)} since they both have the same predicate \texttt{father/2}, and all constants match when the variable $Y$ is grounded to the constant \texttt{Bart}. However, \texttt{dad(Homer,Y)} would not unify since \texttt{dad/2} is not an exact string match with \texttt{father/2}, despite them having the same semantic meaning.

Issues where semantically similar predicates have different wording is common in natural language, so we require a theorem prover which can work with a non-binary unification function based on a similarity score rather than exact string matches.

\begin{figure}
\begin{prooftree}
      \AxiomC{$a_1 \lor \ldots a_n \lor b$}
      \AxiomC{$\neg c \lor d_1 \lor \ldots d_m$}
      \AxiomC{$\textrm{unify}(b,c)$}
      \TrinaryInfC{$a_1 \lor \ldots \lor a_n \lor d_1 \lor \ldots \lor d_m$}
\end{prooftree}
\caption{\label{fig:resolution}Resolution proof rule where each $a_i \in \{a_1,\ldots,a_n\},d_i \in \{d_1,\ldots,d_n\}$ refer to a grounded logical literal, and $b$ and $c$ are positive literals. The condition \texttt{unify(b,c)} holds if the predicates for $b$ and $c$ match, and all constants in all terms also match after grounding variables.}
\end{figure}

The traditional resolution rule \cite{ertel2018introduction} for first-order logic is shown in Figure \ref{fig:resolution}.  A unify function is shown in Algorithm \ref{alg:unifynb}. For simplicity, the algorithm shown assumes that the variable substitution map is provided as input to the unify function, although typically the substitution map is calculated along with the unification. In addition, the algorithm shown does not include function symbols, as those are not used in this paper.

The unify function in Algorithm \ref{alg:unifynb} uses a similarity function, \texttt{simFunc}, which returns a value between 0 and 1, and a threshold $\tau$, where the unification succeeds if the minimum similarity of all similarity checks in the unification is above $\tau$. When \texttt{simFunc} is a string comparison which returns 1 if the strings are identical and 0 if not, and $\tau$ is 0.5, then this reduces to traditional binary unification.

For example, using the example of \texttt{father(Homer, Bart)} and \texttt{father(Homer, Y)} from above, unification succeeds with \texttt{pred1,terms1 = father, [Homer,Bart]}, \texttt{pred2,term2 = father,[Homer,Y]} and \texttt{substitutions = \{Y / Bart\}}.

\begin{algorithm}[ht]
\caption{Non-binary unify}
\label{alg:unifynb}
\textbf{Input}: pred1, terms1 \\
\textbf{Input}: pred2, terms2 \\
\textbf{Input}: substitutions \\
\textbf{Input}: simFunc, $\tau$ \\
\begin{algorithmic} 
\State sim $\gets$ simFunc(pred1, pred2)
\State terms1 $\gets$ applySubs(terms1, substitutions)
\State terms2 $\gets$ applySubs(terms2, substitutions)
\ForAll {term1, term2 $\in$ zip(terms1, terms2)}
    \If {type(term1) $\neq$ type(term2)}
        \Return $False$
    \EndIf
    \If{isConst(term1)}
       \State sim $\gets$ min(sim, simFunc(term1, term2))
    \EndIf
\EndFor
\State \Return sim $>$ $\tau$
\Statex
\Procedure{applySubs}{$terms,subs$}
  \State newTerms $\gets$ []
  \ForAll {term $\in$ terms}
    \State newTerm $\gets$ term
    \If{term $\in$ subs} newTerm $\gets$ subs[term] \EndIf
    \State newTerms $\gets$ newTerms $\cup$ newTerm
  \EndFor
  \State \Return newTerms
\EndProcedure
\end{algorithmic}
\end{algorithm}

We implemented a non-binary unification theorem prover using input resolution called ``Tensor Theorem prover" \footnote{\href{https://github.com/chanind/tensor-theorem-prover}{https://github.com/chanind/tensor-theorem-prover}} which we use in this paper. A proof in Tensor Theorem Prover consists of a chain of resolution steps, each with a corresponding similarity score and substitutions map. The similarity score for the proof as a whole is defined as the minimum similarity of all steps in the proof, as is the case in Neural Theorem Provers \cite{rocktaschel2017end}. Tensor Theorem Prover seeks to find the proof with the highest similarity score for any query, if one exists. We use the unification algorithm presented in the seminal paper on resolution \cite{robinson1965machine} to calculate unification in Tensor Theorem Prover, but replace equality checks between predicates and constants with a non-binary similarity check as shown in Algorithm \ref{alg:unifynb}.


\section{Social reasoning system}
\label{sec:approach}

Our social reasoning system takes a natural language SST and ROT as input. First, we parse the SST and ROT text into token-aligned AMR matched with contextual word embeddings. Next, we generate alternate AMR trees by merging leaf nodes together. From here, we convert AMR trees into first-order logic. Finally, we use a neuro-symbolic theorem prover to query the logical formulae and determine if the ROT is applicable to the situation. Each of these steps will be explained in more detail below.

\begin{figure*}
\centerline{\includegraphics[width=\textwidth]{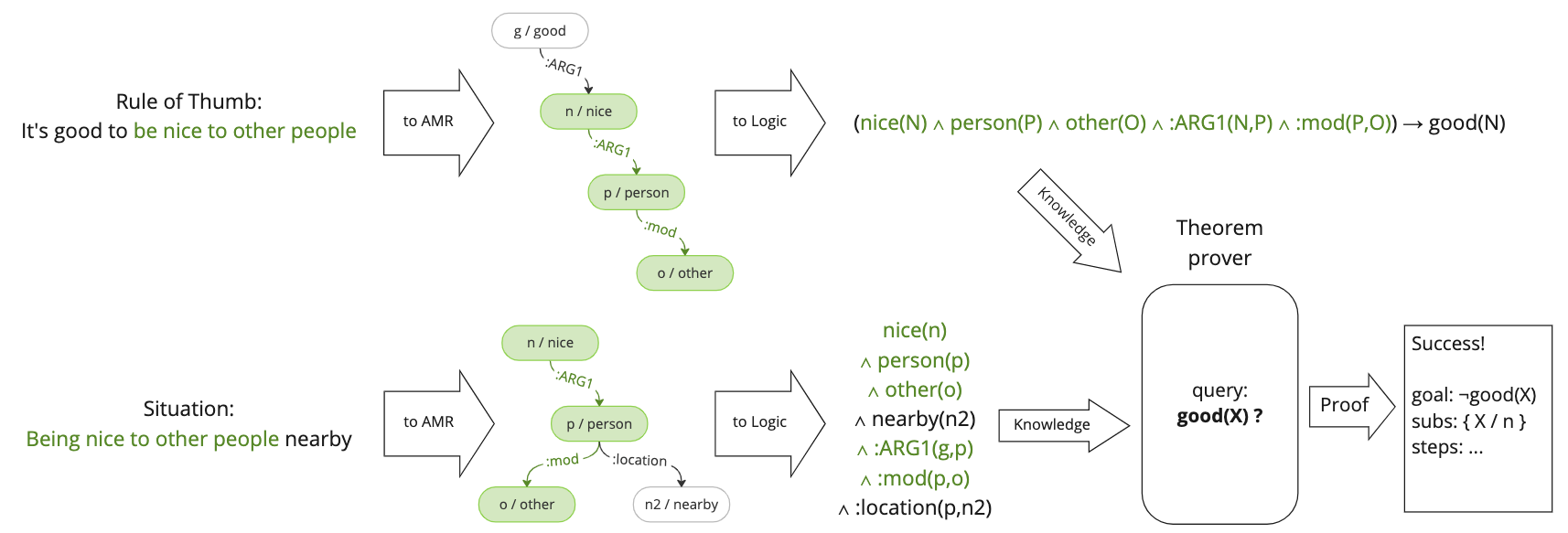}}
\caption{Full flow going from text to AMR to logical formulae to proof for a simple case where the SST AMR and ROT AMR align perfectly with identical predicates and structure. The matching nodes are highlighted in green in the text, AMR, and logical formulae. In formulae, $n$,$p$,$o$,$n2$, and $g$ refer to grounded constants and $N$,$P$, and $O$ refer to universally quantified variables. To address situations where the alignment is not perfect, we introduce the use of contextual embeddings and AMR node merges, described in Section \ref{sec:merge}.}
\label{fig:rotalignbasic}
\end{figure*}

The core insight which allows this to work is that a logical match between a ROT and a situation can be framed as finding a matching subtree in the AMR corresponding to the body of the ROT within the AMR for a situation. This is demonstrated in Figure \ref{fig:rotalignbasic}, where the ROT body AMR subtree is found exactly in the SST AMR. 

\begin{figure}[ht]
\centerline{\includegraphics[width=\linewidth]{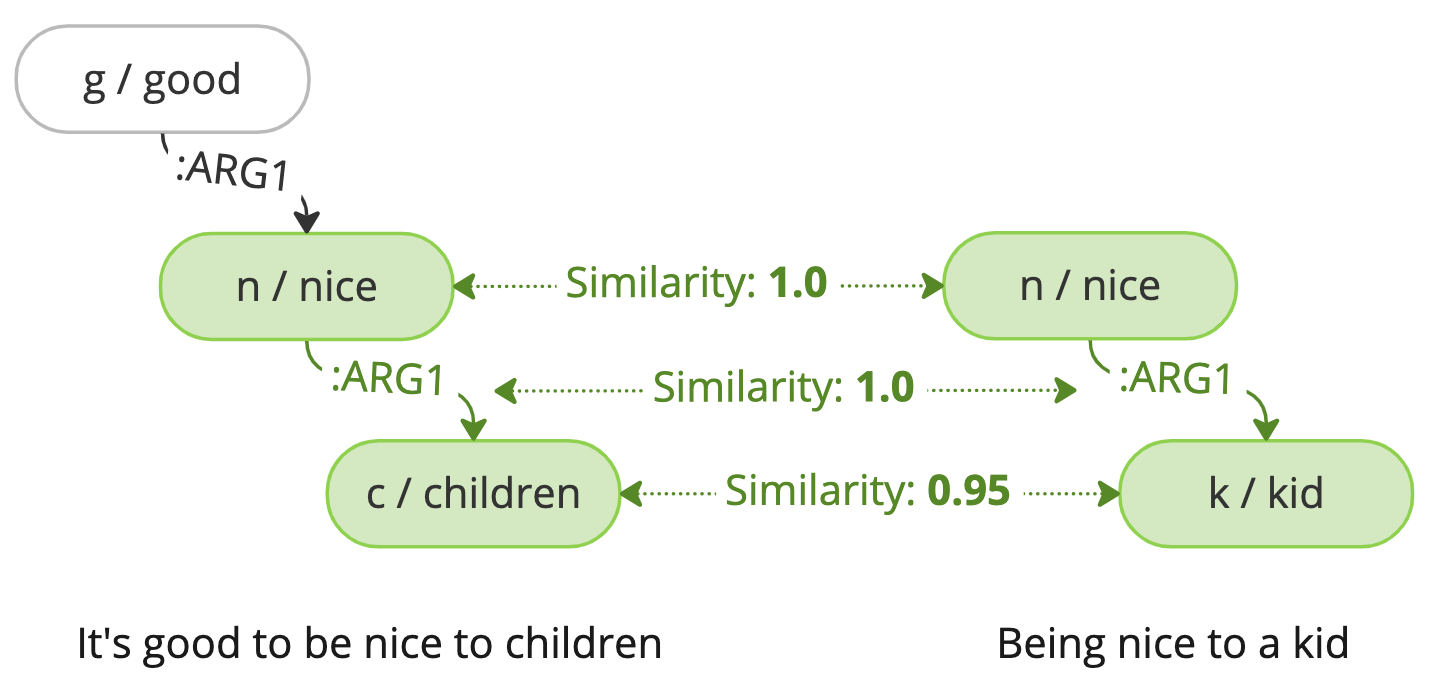}}
\caption{Often nodes in AMR are semantically similar, but not identical. We use contextual embeddings with cosine similarity to allow these nodes to still match. The AMR for the ROT is shown on the left, and the AMR for the SST is shown on the right.}
\label{fig:rotalignembed}
\end{figure}

However, in practice the AMR for the ROT and the SST rarely align so perfectly. To help address this, we introduce contextual embeddings aligned to the nodes in the AMR tree to allow differently worded but semantically similar nodes to still match using a vector similarity score rather than simply performing an exact string match on the AMR instance and role names. This is shown in Figure \ref{fig:rotalignembed}, where the nodes for ``kid" and ``child" are allowed to match due to their embedding vector similarity being very high.

\begin{figure}[ht]
\centerline{\includegraphics[width=\linewidth]{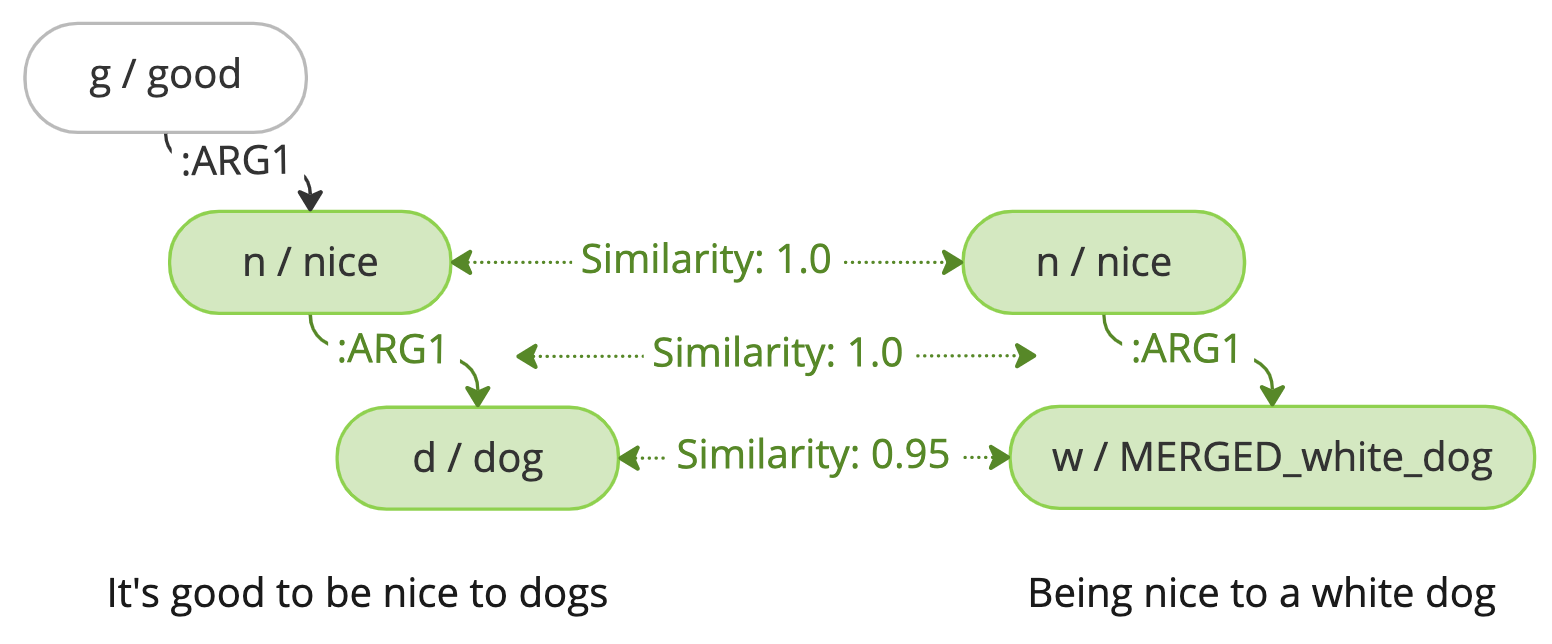}}
\caption{Merging leaf nodes and averaging the merged embeddings allows the ROT (left) and the SST (right) to match, as the resulting tree no longer has a suprious ``w / white-03" node.}
\label{fig:rotalignmerge}
\end{figure}

\subsection{Parsing text into AMR and embeddings}
\label{sec:parse}

The first step when processing ROT and SST text is to generate an AMR tree and contextual embeddings for each sentence. We use a pretrained ensemble AMR 3.0 model for the IBM Transition AMR parser \cite{lee2021maximum} for parsing the ROT and SST into AMR. We use a pretrained RoBERTa \cite{liu2019roberta} base model from Huggingface \cite{wolf2020transformers} and average the last 4 layers of the model to generate contextual embeddings \cite{devlin2018bert}.

An AMR tree with embeddings $\mathcal{T} = \langle \mathcal{N}, \mathcal{E}, meta \rangle$ is a set of nodes $\mathcal{N}$, edges $\mathcal{E}$, and a metadata function $meta$ which provides metadata related to each node and edge. 

The set $\mathcal{N}= \mathcal{N}_I \cup \mathcal{N}_C \cup \mathcal{N}_R$ where 
$\mathcal{N}_I$, $\mathcal{N}_C$, and $\mathcal{N}_R$ are disjoint,
and $\mathcal{N}_I$ consist of AMR instance nodes (of the form \texttt{a / alpha} in Figure \ref{fig:amrtreegraph}), $\mathcal{N}_C$ consist of AMR constants (raw strings like ``Mr Krupp" in Figure \ref{fig:amrlogbos}, numbers, or symbols like ``-" in \texttt{:polarity -}), and $\mathcal{N}_R$ consist of coreference nodes which refer to an instance node (the \texttt{b} coreference in Figure \ref{fig:amrtreegraph}).
Each edge $\langle n, n' \rangle \in \mathcal{E}$ is such that $n \in \mathcal{N}_I$
and $n' \in \mathcal{N}$.
The $meta$ function assigns a tuple of meta-data to each node and edge as follows, where $l$ is a text label, $p$ is a text predicate, and $v$ is a vector embedding from RoBERTa. $v$ can be null (represented as $\emptyset$) for cases where an embedding does not exist.
$$
\begin{array}{ll}
\mbox{If } e \in \mathcal{E}, & meta(e) = \langle l \rangle \nonumber\\
\mbox{If } n \in \mathcal{N}_I, & meta(n) = \langle l, p, v \rangle \nonumber\\
\mbox{If } n \in \mathcal{N}_C, & meta(n) = \langle l, v \rangle \nonumber\\
\mbox{If } n \in \mathcal{N}_R, & meta(n) = \langle l \rangle \nonumber\\
\end{array}
$$

For example, the AMR in Figure \ref{fig:amrlogbos} has 2 instance nodes \texttt{e / dry} with $\langle l = \textrm{e}, p = \textrm{dry}, v = \textrm{embedding[dry]} \rangle$ and \texttt{x / person} with $\langle l = \textrm{x}, p = \textrm{person}, v = \textrm{embedding[``Mr Krupp"]} \rangle$, 1 constant node \texttt{"Mr Krupp"} with $\langle l = \textrm{``Mr Krupp"},  v = \textrm{embedding[``Mr Krupp"]} \rangle$, and 1 coreference node \texttt{x} with $\langle l = \textrm{x} \rangle$. It also contains 3 edges with labels \texttt{:ARG0}, \texttt{:named}, and \texttt{:ARG1}.  

\subsection{Generating merged AMR trees}
\label{sec:merge}

While AMR can smooth over some differences between sentences that have similar semantic meanings, it is still normal for sentences with similar meaning to have slight differences in the structure of the generated AMR tree. When these AMR trees are turned into logical formulae, any differences in the structure of the generated AMR will cause the formula to no longer match even if the sentences are describing semantically similar concepts. Furthermore, the AMR parsing may introduce errors, and we want to be as robust as possible against minor AMR parsing errors.

\begin{figure}[ht]
\centerline{\includegraphics[width=\linewidth]{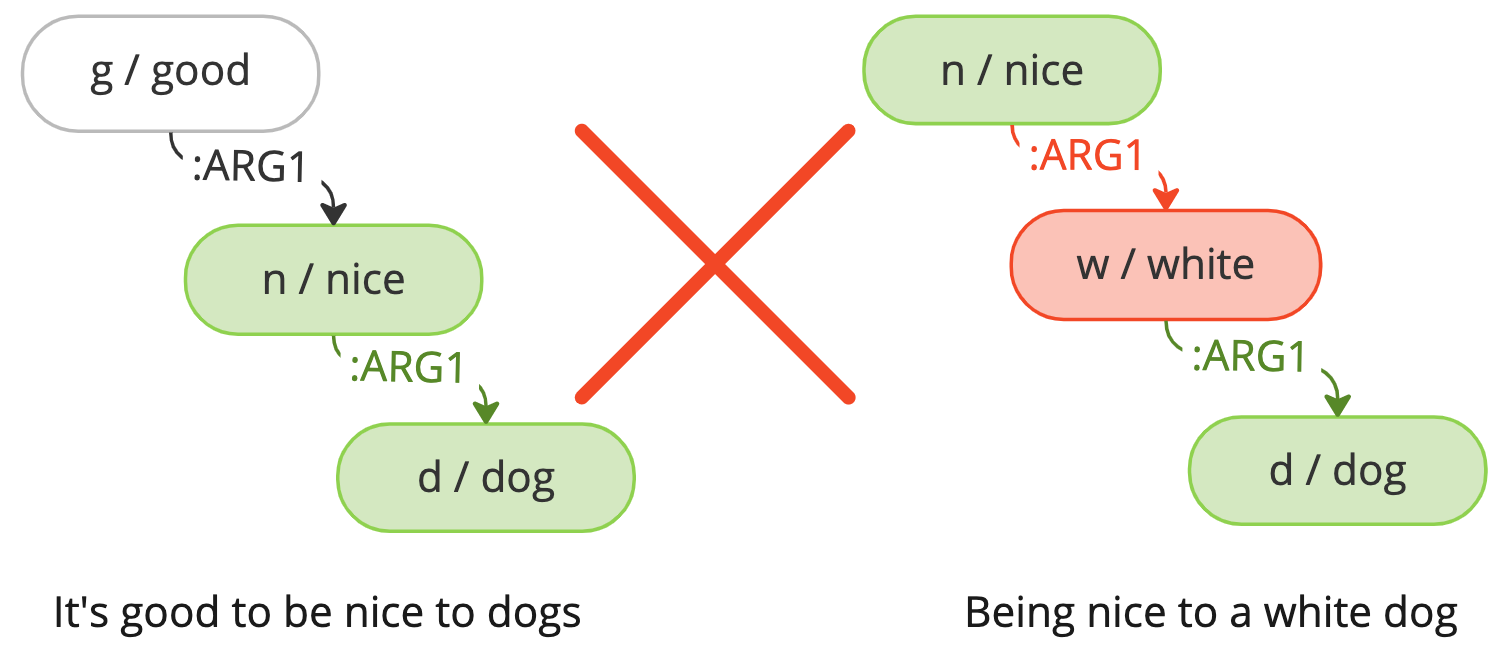}}
\caption{Case where the ROT (left) and SST (right) are logically matching, but the AMR does not align due to an extra node in the SST which is not present in the ROT.}
\label{fig:rotalignmergefail}
\end{figure}

A common failure mode occurs when AMR has a different structure between the ROT and the SST despite the ROT and SST having similar semantic meaning. This manifests itself in extra nodes or missing nodes which break the alignment between ROT and SST. Figure \ref{fig:rotalignmergefail} illustrates this situation, where the text ``white dog" causes an extra node in the AMR tree and breaks alignment with the ROT, which only refers to ``dogs". Of course, a formula that applies to a ``dog" should also apply to a ``white dog".

We address these alignment errors by generating alternative AMR trees by collapsing AMR leaf nodes into a single node, averaging together the embedding vectors of the collapsed nodes. The hope is that the merged embedding vector of the collapsed nodes in the ROT may match the embedding of the corresponding node in the situation via vector similarity, or vice-versa, where without collapsing nodes the AMR graph might have a shape which makes a match impossible. This is illustrated in Figure \ref{fig:rotalignmerge}.


\begin{figure}[ht]
    \centerline{\includegraphics[width=\linewidth]{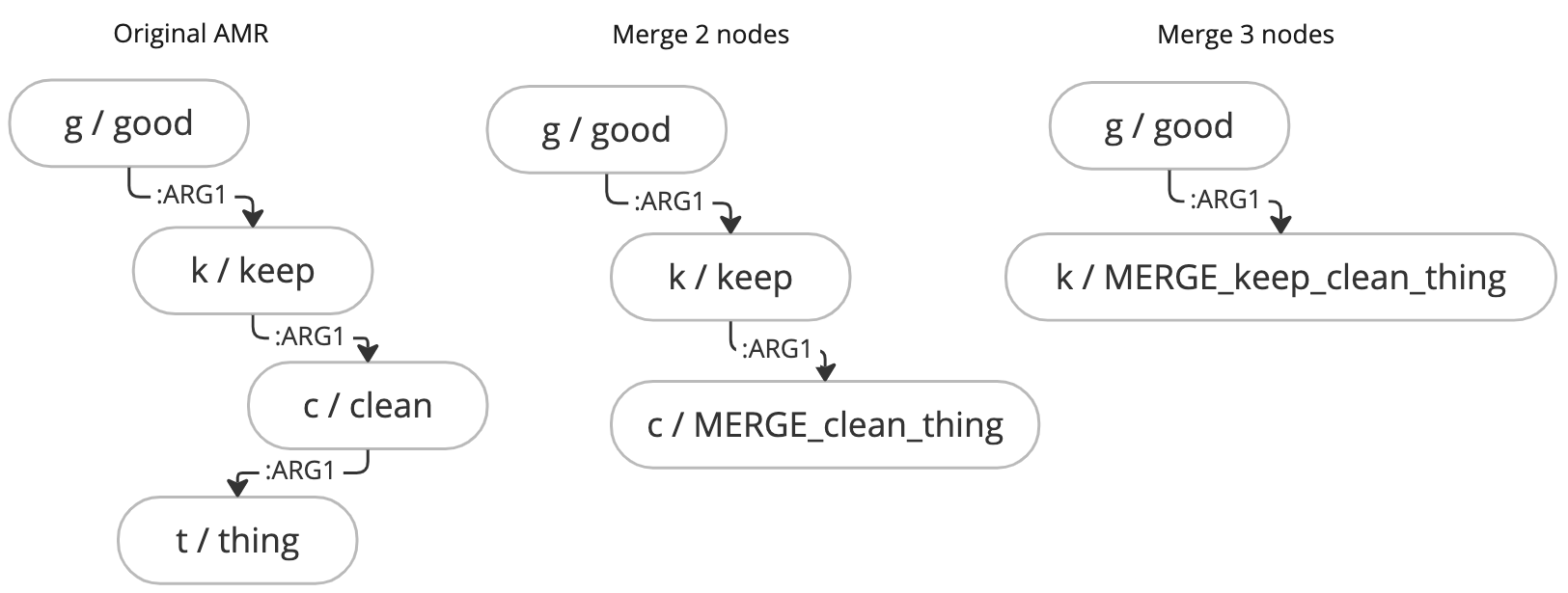}}
    \caption{All possible collapsed AMR trees with a max of 3 merged alignments per node for the AMR for ``It's good to keep things clean"}
    \label{fig:mergedamr}
\end{figure}

We introduce a new node type $\mathcal{N}_M$ to represent a merged node, so that $\mathcal{N} = \mathcal{N}_I \cup \mathcal{N}_C \cup \mathcal{N}_R \cup \mathcal{N}_M$. This node contains a label and the average (using the $avg$ function) of embeddings from the collapsed nodes produced during a merge:
$$
\begin{array}{ll}
    \mbox{If } n \in \mathcal{N}_M, & meta(n) = \langle l, avg(v_1, \ldots v_k) \rangle, k > 0
\end{array}
$$
where $k$, the number of embeddings being merged together, is referred to as the merge width. A node $n'$ is a \emph{child} of a node $n$ if there is an edge from the instance node to the target node. Let $descendent_\mathcal{T}(n, n')$ hold
 if $n$ is a child of $n'$ 
 or if there is an $n''$ s.t. $n$ is a child of $n''$
 and $descendent_\mathcal{T}(n'', n')$ holds. A node $n$ is called the root of tree $\mathcal{T}$,  $n = root_\mathcal{T}$, if there is no edge with $n$ as its target. A node $n$ has depth in $\mathcal{T}$ as defined as follows: 
 $$
 \begin{array}{l}
    depth_\mathcal{T}(n) = 
    \left\{
    \begin{array}{ll}
        0 & \mbox{if } n = root_\mathcal{T}\\
        1 + depth_\mathcal{T}(n') & \mbox{if $\langle n', n \rangle \in \mathcal{E}$}
    \end{array}
    \right.
\end{array}
 $$

Only instance nodes $\mathcal{N}_I$ have children, and thus only these nodes may be collapsed in a merge. To perform a collapse, an instance node $n_I \in \mathcal{N}_I$ is replaced with a merge node $n_M \in \mathcal{N}_M$, such that the label for $n_M$ is the word ``MERGE". The embeddings list for the merge node $n_M$ is a list of all embeddings of $n_I$ and all descendents.

For an AMR tree $\mathcal{T} = \langle \mathcal{N}, \mathcal{E}, meta \rangle$, the AMR tree obtained by merging instance node $n_I$ into new merge node $n_M$ is $\mathcal{T}' = \langle \mathcal{N}', \mathcal{E}', meta' \rangle$  is defined as follows
where 
$J = \{n' | descendent_\mathcal{T}(n',n_I)\} \cup \{n_I\}$, 
$E_m = \{\langle n, n_M \rangle | \langle n, n_I \rangle \in \mathcal{E} \}$,
$l' = ``\mathrm{MERGE}"$,
and $v' = avg(v \in meta(n), v \neq \emptyset | n \in J)$
$$
\begin{array}{l}
    \mathcal{N}' = (\mathcal{N} \setminus J) \cup \{n_M\} \\
    \mathcal{E}' = E_m \cup (\mathcal{E} \cap (\mathcal{N}' \times \mathcal{N}'))\\
    meta'(n)  =
    \left\{
    \begin{array}{ll}
    meta(n) & \mbox{ if } n \in \mathcal{N}\\
    \langle l', v' \rangle & \mbox{ if $n$ is $n_M$}
    \end{array}
    \right.
\end{array}
$$

\begin{figure}[ht]
    \centerline{\includegraphics[width=\linewidth]{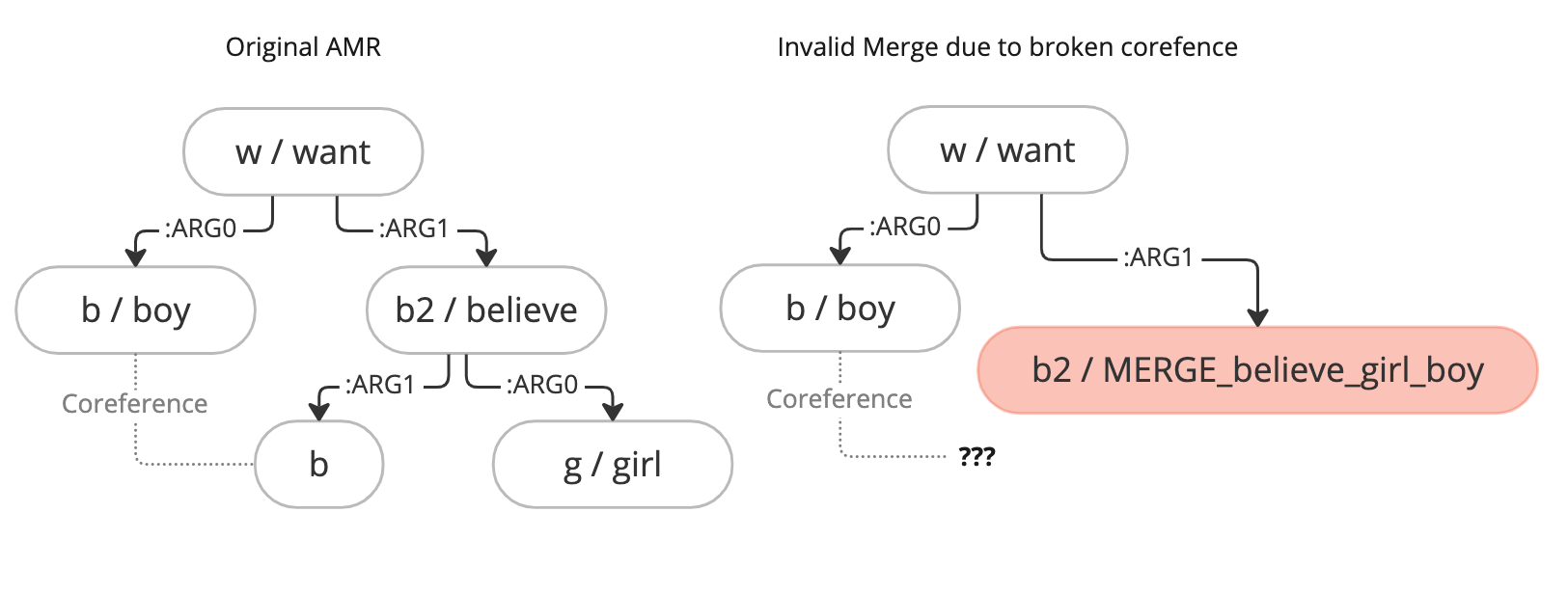}}
    \caption{Invalid merge, since $b$ is coreferenced outside of the collapsed region. The AMR is generated from the text ``The boy wants the girl to believe him".}
    \label{fig:invalidcollapsecoref}
\end{figure}

\begin{figure}[ht]
    \centerline{\includegraphics[width=\linewidth]{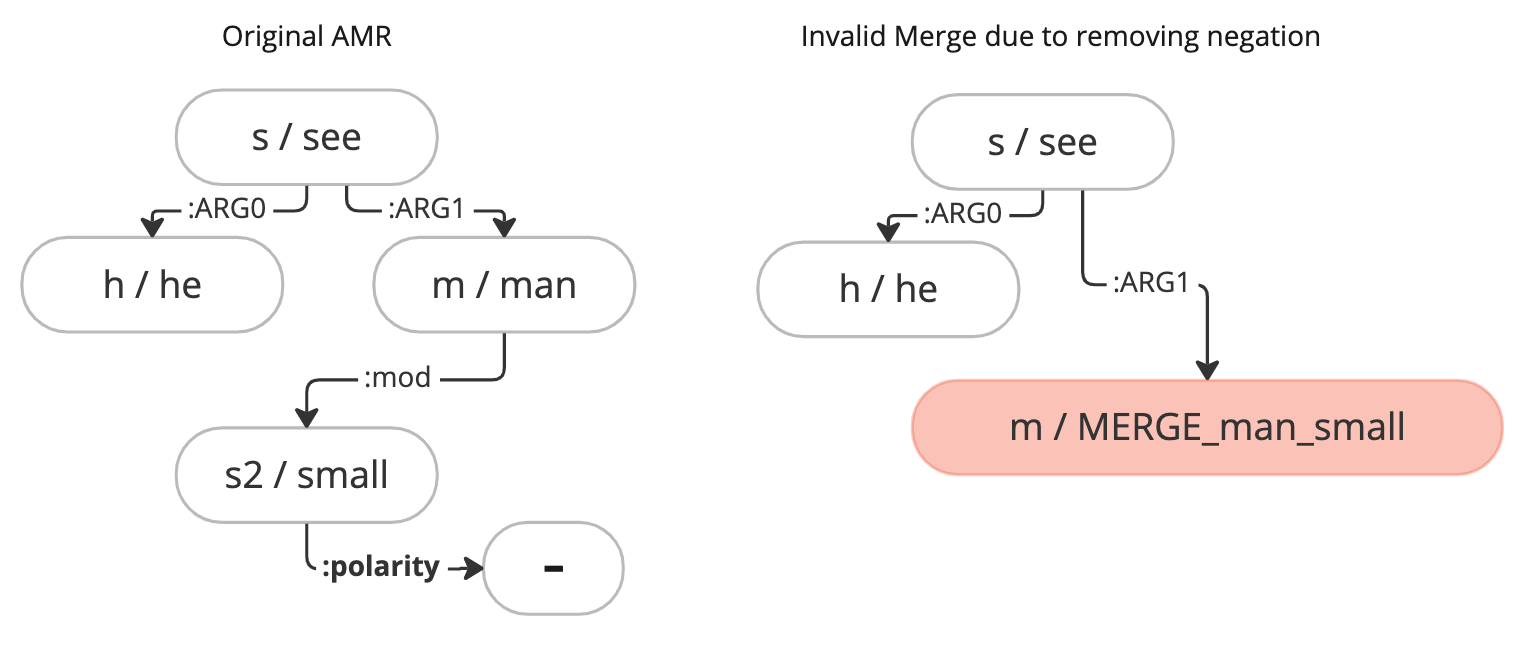}}
    \caption{Invalid merge, since it removes a negation node (:polarity -). The AMR is generated from the text ``He sees a man who is not small"}
    \label{fig:invalidcollapseneg}
\end{figure}

Not all possible merges are considered valid, however. We require that the number of negations (an edge labeled \texttt{:polarity} pointing to a constant node with label \texttt{-}) must be the same in $\mathcal{T}'$ and $\mathcal{T}$. This is demonstrated in Figure \ref{fig:invalidcollapseneg}. Furthermore, if instance node $n_I \in \mathcal{T}$ has a corresponding set of $k$ coreference nodes $\{n_1 \ldots n_k\} \subseteq \mathcal{N}_C, k > 0$, then $\mathcal{T}'$ must either
contain both $n_I$ and all of the corresponding $\{n_1 \ldots n_k\}$ coreferences, or $n_I$ and $\{n_1 \ldots n_k\}$ must be removed. This corresponds to not allowing coreferences to be broken by a merge as this could break the semantic meaning of the AMR, as illustrated in Figure \ref{fig:invalidcollapsecoref}.

We introduce additional parameters for controlling allowed merges futher. $\tau_M$ is a threshold which controls the maximum merge width of the merged embeddings in a single merge node $ meta(n_M) = \langle l, avg(v_1 \ldots v_k) \rangle, k < \tau_M$. $\tau_D$ is a parameter which sets a minimum depth at which merges are allowed in a tree $\mathcal{T}$, so $depth_\mathcal{T}(n_m) > \tau_D$. 

We recursively generate all possible merged AMR trees such that these above conditions are satisfied, including the original tree where no merges have been performed. While this may raise computational complexity concerns if the trees are large, in practice this was never a problem as AMR trees for a single sentence take a negligible amount of time relative to theorem proving and running the AMR parser. Table \ref{tab:statistics} shows statistics on the number of merge trees available per sentence in the Social Chemistry dataset.

\subsection{Converting AMR into logical formulae}

To determine if a Social Chemistry ROT matches a SST, we need to turn both the ROT and SST AMR into first-order logical formulae. Until now, we have treated both the ROT and the SST identically, but when translating the ROT and SST into logical formulae we need to treat them differently.

A Social Chemistry ROT is a general logical rule which can apply to any number of situations. A ROT typically has the form of an action and a verdict on that action. For instance, the ROT ``It's rude to hang up on someone." would have the verdict of ``It's rude", and action ``to hang up on someone".

We use this knowledge of how ROTs are structured to modify the AMR to formulae conversion so the result is in the form of an implication. We start with the AMR to logical formulae algorithm described by \cite{bos2016expressive}, but we make modifications to the quantification and the structure of the formulae. The \cite{bos2016expressive} algorithm wraps all instances with existential quantifiers, which assumes that each sentence is a statement which can be true or false. However, natural language is able to express much more than this, and we use our knowledge of what is being expressed in the ROT and the SST to justify changing the formulae. For a ROT, the formulae corresponds to an implication as mentioned above. For a SST, the formulae describes grounded knowledge which we can query to see if the ROT matches.

For instance, we can begin with the ROT AMR tree below for ``It's rude to hang up on someone.":

\begin{footnotesize}
\begin{verbatim}
(r / rude
    :ARG1 (h / hang-up
        :ARG2 (s / someone)))
\end{verbatim}
\end{footnotesize}

From here, the algorithm from \cite{bos2016expressive} would result in the following formulae:
\begin{align}
    & \exists R.(\textrm{rude}(R) \land \exists H.(\textrm{:ARG1}(R,H) \land  \textrm{hang-up}(H) \nonumber\\
    & \hspace{5mm} \land \; \exists S.(\textrm{:ARG2}(H,S) \land  \textrm{someone}(S)))) \nonumber
\end{align}

We know that the ROT is a logical implication, so we remove the existential quantification. Then, we replace the conjunction between the verdict, $\textrm{rude}(R)$ with implication, and remove the linking $\textrm{:ARG1}(R,H)$, since this linking is handled by the implication itself. In addition, we swap the $R$ variable in  $\textrm{rude}(R)$ with $H$, as this is the target of $\textrm{:ARG1}(R,H)$. This results in the following implication:
\begin{align}
    \textrm{hang-up}(H) \land \textrm{:ARG2(H,S)} \land \textrm{someone}(S) \to \textrm{rude}(H) \nonumber
\end{align}
As a final step, $\textrm{rude}(H)$ is simplified further to just $\textrm{BAD}(H)$, resulting in the following final logical form:
\begin{align}
    \textrm{hang-up}(H) \land \textrm{:ARG2(H,S)} \land  \textrm{someone}(S) \to \textrm{BAD}(H) \nonumber
\end{align}

Converting SST AMR to logical formulae is much simpler, as there is no need to build an implication. Instead, the only changes necessary from the base algorithm from \cite{bos2016expressive} is to remove existential quantifiers, and to replace variables with constants. Constants are used to indicate that the formulae represents facts about items in the domain of discourse, rather than general theorems. For instance, we can begin with the SST AMR tree below for ``Hanging up on my cousin":

\begin{footnotesize}
\begin{verbatim}
(h / hanging
    :ARG2 (p / person
        :ARG0-of (h2 / have-rel-role
            :ARG1 (i / i)
            :ARG2 (c / cousin))))
\end{verbatim}
\end{footnotesize}

Following \cite{bos2016expressive}, this becomes the following logic:

\begin{align}
    & \exists H.(\textrm{hanging}(H) \land \exists P.(\textrm{:ARG2}(H, P) \nonumber\\
    & \hspace{5mm} \land \; \textrm{person}(P) \land \exists H2.(\textrm{:ARG0}(H2, P) \nonumber\\
    & \hspace{5mm} \land \; \textrm{have-rel-role}(H2) \land \exists I.(\textrm{:ARG1}(H2, I) \nonumber\\
    & \hspace{5mm} \land \; \textrm{i}(I)) \land \exists C.(\textrm{:ARG2}(H2, C) \land \textrm{cousin}(C))))) \nonumber
\end{align}

From here, we ground out the existential quantifiers and replace variables with new constants, resulting in the following final logical form. Note that in these examples, lowercase letters indicate a constant, and uppercase letters indicate a variable.

\begin{align}
    & \mbox{hanging}(h) \land \mbox{:ARG2}(h, p) \land \mbox{person}(p) \nonumber\\
    & \hspace{5mm} \land \; \mbox{:ARG0}(h2, p) \land \mbox{have-rel-role}(h2) \land \mbox{:ARG1}(h2, i) \nonumber\\
    & \hspace{5mm} \land \; \mbox{i}(i) \land \mbox{:ARG2}(h2, i) \land \mbox{cousin}(i) \nonumber
\end{align}

While the above process is shown only for the original AMR tree, this process is also repeated for all merged AMR tree versions of the ROT and SST.

\subsection{Reasoning}

Once logical formulae are generated for a ROT and a SST, we use the Tensor theorem prover library discussed in Section \ref{sec:theoremproving} to check if the SST applies to the antecedent of the ROT. Since each ROT is transformed into an implication with a consequent of either $\textrm{GOOD}(X)$, $\neg \textrm{GOOD}(X)$, $\textrm{BAD}(X)$, or $\neg \textrm{BAD}(X)$, we can verify if the ROT matches by querying these 4 verdicts one by one in the theorem prover. If a proof can be found for any of these queries, then the ROT matches the SST.

For unification, we use a hybrid similarity function which can perform exact string match or cosine similarity between embeddings, depending on the predicates or constants being compared. This is shown in Algorithm \ref{alg:simalgo}, and is used as the \texttt{simFunc} parameter in unification in Algorithm \ref{alg:unifynb}. Here, symbol1 and symbol2 are either constants or predicates, and embedding1 and embedding2 are their RoBERTa embeddings, but may be null.

\begin{algorithm}[ht]
\caption{Hybrid Similarity Function}
\label{alg:simalgo}
\textbf{Input}: symbol1, embedding1 \\
\textbf{Input}: symbol2, embedding2 \\
\textbf{Output}: similarity
\begin{algorithmic} 

\State bothHaveEmbeds $\gets$ embedding1 $\land$ embedding2
\State eitherIsMerge $\gets$ ``MERGED" $\in$ [symbol1, symbol2]
\If {eitherIsMerge}
    \If {bothHaveEmbeds}
        \State \Return $\frac{1}{2} (cos(\textrm{embedding1}, \textrm{embedding2}) + 1)$
    \Else
        \State \Return 0.0
    \EndIf
\EndIf
\If {symbol1 $=$ symbol2}
    \State \Return 1.0
\EndIf
\If {bothHaveEmbeds}
    \State \Return $\frac{1}{2} (cos(\textrm{embedding1}, \textrm{embedding2}) + 1)$
\EndIf
\State \Return 0.0

\end{algorithmic}
\end{algorithm}

We treat the minimum similarity threshold above which a unification is allowed to succeed as a tunable parameter of the theorem proving process, but find that using a value near 0.9 performs well.

\section{Experimental evaluation}
\label{sec:eval}

To evaluate the performance of this approach, we use the SST and ROT from Social Chemistry 101 as a test bed. For each sample, we turn the ROT and SST into logical formulae as described in Section \ref{sec:approach}, and check if the the theorem prover is able to prove the verdict of the ROT from the SST. To create negative examples for evaluation, we randomly select a different ROT from the dataset and check that the theorem prover should NOT be able to prove the conclusion of that ROT when applied to the original SST. Based on these, we calculate precision, recall, and f1 score.

For example, given the SST ``being friends with my ex's sister", and the corresponding ROT ``You shouldn't be friend with your ex's family members", the ROT formulae has the following form:
\begin{align}
    & (\textrm{you}(Y) \land \textrm{:ARG1}(H,Y) \land \textrm{have-rel-role}(H) \ldots) \nonumber\\
    & \hspace{5mm} \to \neg \textrm{GOOD}(Y) \nonumber
\end{align}
Since the consequent of this ROT is $\neg \textrm{GOOD}(Y)$, if we can prove $\neg \textrm{GOOD}(Y)$ using Tensor theorem prover after inputting the logical formulae for the SST and the ROT, then this is considered a true positive. Likewise, if we cannot find a proof of $\neg \textrm{GOOD}(Y)$ then this a false negative.

We also pick an unrelated ROT, and correspondingly see if we can prove the consequent of that ROT given the SST. For instance, if we randomly pick the unrelated ROT ``You should always give gifts to people for their birthday", and we are able to find a proof for the consequent of this ROT, $\textrm{GOOD}(X)$, when applied to the unrelated SST ``being friends with my ex's sister", then this is considered a false positive. Likewise, if we are unable to prove the consequent of the unrelated ROT, then this is considered a true negative.

We randomly selected 10,000 samples consisting of a ROT and corresponding SST from the social chemistry 101 dataset as a test set. Statistics related to the AMR trees for the SSTs and ROTs in this dataset are shown in Table \ref{tab:selectedrots}.

\begin{table}[h!]
    \begin{center}
        {\renewcommand{\arraystretch}{1.2}%
        \begin{tabular}{ |p{3cm}||p{1.2cm}|p{1.2cm}|p{1.2cm}|  }
         \hline
         \multicolumn{4}{|c|}{Dataset statistics} \\
         \hline
         \thead{ROT AMR} & \thead{Mean} & \thead{Median} & \thead{Stdev}\\
         \hline
         Instance nodes    & 7.1  & 7  & 2.5\\
         AMR depth         & 5.2  & 5  & 1.3\\
         Logic terms       & 13.6 & 13 & 5.9\\
         Merge Trees       & 1.6  & 1  & 2.2\\
         \hline
         \thead{SST AMR} & \thead{Mean} & \thead{Median} & \thead{Stdev}\\
         \hline
         Instance nodes    & 8.6  & 8  & 3.1\\
         AMR depth         & 5.1  & 5  & 1.4\\
         Logic terms       & 18.4 & 17 & 9.1\\
         Merge Trees       & 2.4  & 1  & 14.1\\
         \hline
        \end{tabular}}
        \vskip3mm
        \caption{Statistics for the ROT AMR and SST AMR generated from the Social Chemistry 101 dataset. Instance nodes refers to AMR instances (things like b/boy in AMR), and logic terms is the number of literal in our generated logical formulae before merging. Merge trees is the number of alterate merge trees that can be generated from a SST or ROT AMR tree}
        \label{tab:statistics}
    \end{center}
\end{table}

\begin{figure}[ht]
\centerline{\includegraphics[width=\linewidth]{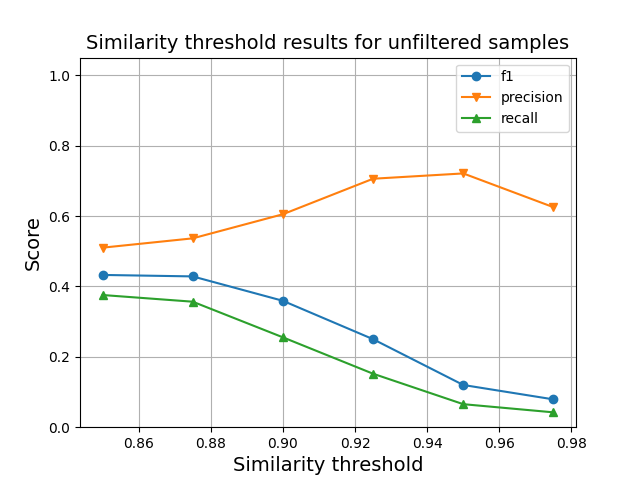}}
\caption{Precision, Recall, and F1 score, varying the similarity threshold}
\label{fig:unfilteredscoresthresh}
\end{figure}

Our method includes a hyperparameter for the vector similarity threshold at which a unification is considered a success in the theorem prover. Raising this threshold will increase precision but at the cost of recall. The precision, recall, and f1 score as a function of similarity threshold is shown for in Figure \ref{fig:unfilteredscoresthresh}. We set a minimum merge depth of 1, and a maximum leaf merge width of 6 nodes.

\begin{figure}[ht]
\centerline{\includegraphics[width=\linewidth]{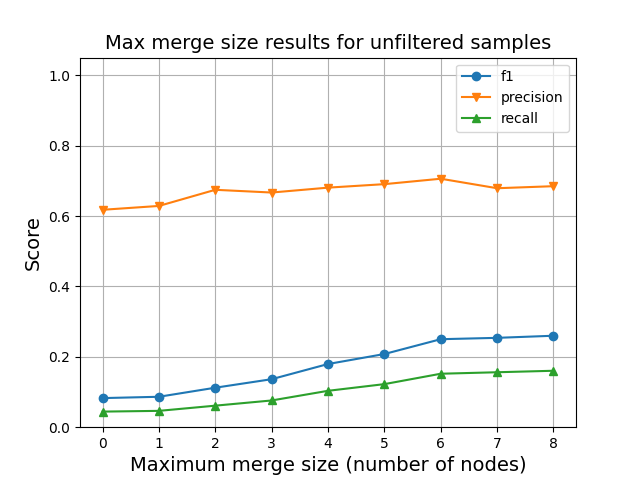}}
\caption{Precision, Recall, and F1 score, varying the maximum merge size}
\label{fig:unfilteredscoresmerge}
\end{figure}

We also investigated the effect of the number of merged nodes on the performance of this approach. For this experiment, we set the similarity threshold to 0.925, and kept the remaining parameters identical to the similarity threshold experiment. The results are shown in Figure \ref{fig:unfilteredscoresmerge}.

Increasing the allowed number of nodes to be merged does not seem to have a large effect on precision, but it does have a large impact on recall and f1 score up to a merge size of 6 nodes. Increasing the max node merge size allows the prover to rely more on the embeddings to unify large chunks of the graph rather than relying on the AMR structure, so it is encouraging to see these merges don't seem to have a negative effect on precision up to 6 nodes.

The merge algorithm we propose has a number of restrictions on when merges are possible. The most common causes of merges not being allowed come from crossing negation bounds, and merging across coreferences. However, these restrictions also limit the performance of our method overall. We introduce a metric called collapsability, which measures what portion of an AMR tree can be collapsed following our merge algorithm. For a set of AMR trees $\{\mathcal{T}_1 \dots \mathcal{T}_n\}$, minNodes($\{\mathcal{T}_1 \dots \mathcal{T}_n\}$) and maxNodes($\{\mathcal{T}_1 \dots \mathcal{T}_n\}$) refer to the number of nodes in the smallest and largest trees in the set, respectively. Collapsability is thus defined below for an AMR tree $\mathcal{T}$:
$$
\textrm{collapsability}(\mathcal{T}) = 1 - \frac{\textrm{minNodes}(\textrm{merges}(\mathcal{T})) - 1}{\textrm{maxNodes}(\textrm{merges}(\mathcal{T})) - 1}
$$

Collapsability is 0 if no merges are possible, and 1 if the entire AMR tree can be collapsed into a single merged node. Collapsability is undefined if the original AMR tree is just 1 node as that would cause $maxNodes$ to be 1, but this never occurs in our dataset.

\begin{figure}[ht]
\centerline{\includegraphics[width=\linewidth]{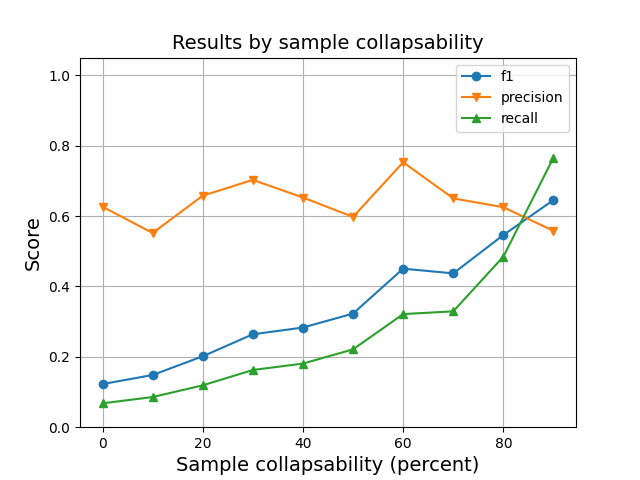}}
\caption{Precision, Recall, and F1 score when bucketing samples based on what percentage of the AMR is allowed to be collapsed (merged)}
\label{fig:collapsability}
\end{figure}

Figure \ref{fig:collapsability} shows the results when considering only subsets of the dataset bucketed by the collapsability percent of the ROT and the SST. The minimum collapsability between these is used as the collapsability of the sample.

In these results, we see the importance of allowing merges on the performance of our method. When few merges are possible, the recall is very low, since the majority of the AMR for ROT and SST do not have identical structure. As the collapsability of samples increases, the recall increases dramatically as well, even surpassing the precision at around 80\%+ collapsability. Interestingly, at very high collapsability the precision declines slightly, likely since when so many nodes are merged together it is easier to get false positives due to excessive averaging of embeddings. Regardless, the performance increase from higher collapsability should motivate future work to improve the merge algorithm to allow merging across coreference and negation.

\section{Related work}
\label{sec:relatedwork}


The idea of using a semantic parser to turn natural language into a logical format for reasoning has been used in \cite{sharma2015towards,hong2022karaml} for solving Winograd schema challenges (WSCs) \cite{levesque2012winograd}. Both of these papers use a semantic parser called K-Parser \cite{sharma2015identifying} to parse WSC sentences and turn them into a logical form that can be reasoned with. \cite{sharma2015towards} then tries to find a theorem which will complete the WSC and uses Answer set programming (ASP) \cite{gelfondstable} with the semantic graph output from K-Parser to complete the WSC task. K-Parser is conceptually similar to AMR, but is missing key features of AMR like negation and standardized roles, and does not appear to be actively maintained. This approach does not use vector embeddings or other neuro-symbolic approaches to improve the robustness of matches, and thus cannot handle formulae which do not match the exact structured output of K-Parser exactly.

There has been work done on relaxing the unification condition of theorem provers to allow for differentiable unification, from which our Tensor Theorem Prover library takes inspiration. Neural Theorem Provers (NTPs) \cite{rocktaschel2017end} and Braid \cite{kalyanpur2022braid} allow unification to return a score between 0 and 1. This is typically a similarity score between vector embeddings corresponding to each predicate. However, both of these libraries are back-chaining provers, and can only work with Horn clauses which is insufficient to represent the full range of formulae present in Social Chemistry ROTs. Also conceptually similar is \cite{hunter2022understanding}, which allows using embedding vector similarity to swap logical predicates using a SAT solver.

The Social Chemistry 101 dataset includes a number of extra dimensions to the data relating to ethics, psychology, and morality around samples in the dataset. Trained models on this dataset tend to focus on generating judgements conditioned on these attributes \cite{forbes2020social}, or using the data as a component in other more upstream datasets such as defeasible NLI \cite{rudinger2020thinking} or focusing on the intentions of actors in social situations \cite{emelin2020moral}. In all these cases, the dataset is used for different purposes than matching ROTs with SSTs as is done in this paper. 

Perhaps most similar to our paper is \cite{kapanipathi2020leveraging}. Here, a natural-language query is turned into AMR, and the AMR is turned into logical formulae. However, these formulae are then used to query a pre-existing knowledge-base using SPARQL rather than deriving the knowledge-base as well from a natural language sentence. Furthermore, the focus is on entity recognition and data retrieval rather than social commonsense reasoning, and does not use contextual word embedding similarity for neuro-symbolic reasoning.

\section{Discussion}
\label{sec:discussion}

 In this paper, we present a novel system for taking ROTs in natural language and reasoning with them using a neuro-symbolic theorem prover. Our contributions in this paper include: (1) an approach for merging and collapsing AMR nodes for increased flexibility and robustness during reasoning; (2) a hybrid similarity metric which mixes string matching and embedding cosine similarity; (3) a modified version of AMR to logic conversion for working with ROT implications; and (4) our evaluation method using Social Chemistry 101. Furthermore, this paper introduces Tensor Theorem Prover, our implementation of a resolution-based neuro-symbolic theorem prover and AMR Logic Converter, our library for converting AMR to first-order logic.

While this paper deals only with a single SST and ROT at a time, this can be extended in the future to allow multi-hop reasoning by combining multiple ROTs and additional background knowledge during the reasoning process.

The theorem proving process in this paper is differentiable. In future work the embeddings used for semantic similarity can be further trained via backpropagation.

A particular weakness of the current approach is that it does not have a good way to deal with antonyms during the solving process. For example, if a ROT references ``gifts that aren't too big", it would be reasonable that a SST describing a "small gift" to match, since ``big" and ``small" are opposites. A formula like $\textrm{big}(X) \to \neg \textrm{small}(X)$ could be injected dynamically to address this.


Another area in which this approach could be improved is dealing with multiple sentences. This would require working out coreferences between entities in the various sentences, but doing so should allows this approach to work with paragraphs of text instead of single sentences.

A final area of future work is be to develop a version of the merge algorithm which can allow merging nodes across coreferences and/or negation. Our evaluation shows better results when more merging is possible, so a clear direction for improvement is to develop a version of the merge algorithm which has fewer restrictions on allowed merges.

The method described in this paper is a stepping-stone on the path to building systems capable of social reasoning using an interpretable, neuro-symbolic approach. Systems like this will naturally be useful anywhere commonsense social understanding is necessary, from interacting with users directly, to summarizing human narrative, to giving recommendations on social situations. In all these cases, and especially when dealing with sensitive social topics, it is important for future AI systems to be interpretable and transparent, and for their reasoning to be debuggable and editable.

\bibliographystyle{kr}  
\bibliography{references}

\begin{thebibliography}{}

\bibitem[\protect\citeauthoryear{Abid, Farooqi, and
  Zou}{2021}]{abid2021persistent}
Abid, A.; Farooqi, M.; and Zou, J.
\newblock 2021.
\newblock Persistent anti-muslim bias in large language models.
\newblock In {\em Proceedings of the 2021 AAAI/ACM Conference on AI, Ethics,
  and Society},  298--306.

\bibitem[\protect\citeauthoryear{Banarescu \bgroup et al\mbox.\egroup
  }{2013}]{banarescu2013abstract}
Banarescu, L.; Bonial, C.; Cai, S.; Georgescu, M.; Griffitt, K.; Hermjakob, U.;
  Knight, K.; Koehn, P.; Palmer, M.; and Schneider, N.
\newblock 2013.
\newblock Abstract meaning representation for sembanking.
\newblock In {\em Proceedings of the 7th linguistic annotation workshop and
  interoperability with discourse},  178--186.

\bibitem[\protect\citeauthoryear{Bolukbasi \bgroup et al\mbox.\egroup
  }{2016}]{bolukbasi2016man}
Bolukbasi, T.; Chang, K.-W.; Zou, J.~Y.; Saligrama, V.; and Kalai, A.~T.
\newblock 2016.
\newblock Man is to computer programmer as woman is to homemaker? debiasing
  word embeddings.
\newblock {\em Advances in neural information processing systems} 29.

\bibitem[\protect\citeauthoryear{Bos}{2016}]{bos2016expressive}
Bos, J.
\newblock 2016.
\newblock Expressive power of abstract meaning representations.
\newblock {\em Computational Linguistics} 42(3):527--535.

\bibitem[\protect\citeauthoryear{Devlin \bgroup et al\mbox.\egroup
  }{2018}]{devlin2018bert}
Devlin, J.; Chang, M.-W.; Lee, K.; and Toutanova, K.
\newblock 2018.
\newblock Bert: Pre-training of deep bidirectional transformers for language
  understanding.
\newblock {\em arXiv preprint arXiv:1810.04805}.

\bibitem[\protect\citeauthoryear{Drozdov \bgroup et al\mbox.\egroup
  }{2022}]{drozdov2022inducing}
Drozdov, A.; Zhou, J.; Florian, R.; McCallum, A.; Naseem, T.; Kim, Y.; and
  Astudillo, R.~F.
\newblock 2022.
\newblock Inducing and using alignments for transition-based amr parsing.
\newblock {\em arXiv preprint arXiv:2205.01464}.

\bibitem[\protect\citeauthoryear{Emelin \bgroup et al\mbox.\egroup
  }{2020}]{emelin2020moral}
Emelin, D.; Bras, R.~L.; Hwang, J.~D.; Forbes, M.; and Choi, Y.
\newblock 2020.
\newblock Moral stories: Situated reasoning about norms, intents, actions, and
  their consequences.
\newblock {\em arXiv preprint arXiv:2012.15738}.

\bibitem[\protect\citeauthoryear{Ertel}{2018}]{ertel2018introduction}
Ertel, W.
\newblock 2018.
\newblock {\em Introduction to artificial intelligence}.
\newblock Springer.

\bibitem[\protect\citeauthoryear{Forbes \bgroup et al\mbox.\egroup
  }{2020}]{forbes2020social}
Forbes, M.; Hwang, J.~D.; Shwartz, V.; Sap, M.; and Choi, Y.
\newblock 2020.
\newblock Social chemistry 101: Learning to reason about social and moral
  norms.
\newblock {\em arXiv preprint arXiv:2011.00620}.

\bibitem[\protect\citeauthoryear{Gelfond and Lifschitz}{}]{gelfondstable}
Gelfond, M., and Lifschitz, V.
\newblock The stable model semantics for logic programming. icslp, 1988.

\bibitem[\protect\citeauthoryear{Goodman}{2020}]{goodman-2020-penman}
Goodman, M.~W.
\newblock 2020.
\newblock {P}enman: An open-source library and tool for {AMR} graphs.
\newblock In {\em Proceedings of the 58th Annual Meeting of the Association for
  Computational Linguistics: System Demonstrations},  312--319.
\newblock Online: Association for Computational Linguistics.

\bibitem[\protect\citeauthoryear{Hong \bgroup et al\mbox.\egroup
  }{2022}]{hong2022karaml}
Hong, S.~J.; Bennett, B.; Clymo, J.; and {\'A}lvarez, L.~G.
\newblock 2022.
\newblock Karaml: Integrating knowledge-based and machine learning approaches
  to solve the winograd schema challenge.
\newblock In {\em AAAI Spring Symposium: MAKE}.

\bibitem[\protect\citeauthoryear{Hovy \bgroup et al\mbox.\egroup
  }{2006}]{hovy2006ontonotes}
Hovy, E.; Marcus, M.; Palmer, M.; Ramshaw, L.; and Weischedel, R.
\newblock 2006.
\newblock Ontonotes: the 90\% solution.
\newblock In {\em Proceedings of the human language technology conference of
  the NAACL, Companion Volume: Short Papers},  57--60.

\bibitem[\protect\citeauthoryear{Hunter}{2022}]{hunter2022understanding}
Hunter, A.
\newblock 2022.
\newblock Understanding enthymemes in deductive argumentation using semantic
  distance measures.
\newblock In {\em Proceedings of the AAAI Conference on Artificial
  Intelligence}, volume~36,  5729--5736.

\bibitem[\protect\citeauthoryear{Kalyanpur, Breloff, and
  Ferrucci}{2022}]{kalyanpur2022braid}
Kalyanpur, A.; Breloff, T.; and Ferrucci, D.~A.
\newblock 2022.
\newblock Braid: Weaving symbolic and neural knowledge into coherent logical
  explanations.
\newblock In {\em Proceedings of the AAAI Conference on Artificial
  Intelligence}, volume~36,  10867--10874.

\bibitem[\protect\citeauthoryear{Kapanipathi \bgroup et al\mbox.\egroup
  }{2020}]{kapanipathi2020leveraging}
Kapanipathi, P.; Abdelaziz, I.; Ravishankar, S.; Roukos, S.; Gray, A.;
  Astudillo, R.; Chang, M.; Cornelio, C.; Dana, S.; Fokoue, A.; et~al.
\newblock 2020.
\newblock Leveraging abstract meaning representation for knowledge base
  question answering.
\newblock {\em arXiv preprint arXiv:2012.01707}.

\bibitem[\protect\citeauthoryear{Kim \bgroup et al\mbox.\egroup
  }{2018}]{kim2018interpretability}
Kim, B.; Wattenberg, M.; Gilmer, J.; Cai, C.; Wexler, J.; Viegas, F.; et~al.
\newblock 2018.
\newblock Interpretability beyond feature attribution: Quantitative testing
  with concept activation vectors (tcav).
\newblock In {\em International conference on machine learning},  2668--2677.
\newblock PMLR.

\bibitem[\protect\citeauthoryear{Kingsbury and
  Palmer}{2002}]{kingsbury2002treebank}
Kingsbury, P.~R., and Palmer, M.
\newblock 2002.
\newblock From treebank to propbank.
\newblock In {\em LREC},  1989--1993.

\bibitem[\protect\citeauthoryear{Lee \bgroup et al\mbox.\egroup
  }{2021}]{lee2021maximum}
Lee, Y.-S.; Astudillo, R.~F.; Hoang, T.~L.; Naseem, T.; Florian, R.; and
  Roukos, S.
\newblock 2021.
\newblock Maximum bayes smatch ensemble distillation for amr parsing.
\newblock {\em arXiv preprint arXiv:2112.07790}.

\bibitem[\protect\citeauthoryear{Levesque, Davis, and
  Morgenstern}{2012}]{levesque2012winograd}
Levesque, H.; Davis, E.; and Morgenstern, L.
\newblock 2012.
\newblock The winograd schema challenge.
\newblock In {\em Thirteenth international conference on the principles of
  knowledge representation and reasoning}.

\bibitem[\protect\citeauthoryear{Liu \bgroup et al\mbox.\egroup
  }{2019}]{liu2019roberta}
Liu, Y.; Ott, M.; Goyal, N.; Du, J.; Joshi, M.; Chen, D.; Levy, O.; Lewis, M.;
  Zettlemoyer, L.; and Stoyanov, V.
\newblock 2019.
\newblock Roberta: A robustly optimized bert pretraining approach.
\newblock {\em arXiv preprint arXiv:1907.11692}.

\bibitem[\protect\citeauthoryear{Lu \bgroup et al\mbox.\egroup
  }{2020}]{lu2020gender}
Lu, K.; Mardziel, P.; Wu, F.; Amancharla, P.; and Datta, A.
\newblock 2020.
\newblock Gender bias in neural natural language processing.
\newblock {\em Logic, Language, and Security: Essays Dedicated to Andre Scedrov
  on the Occasion of His 65th Birthday}  189--202.

\bibitem[\protect\citeauthoryear{Lundberg and Lee}{2017}]{NIPS2017_7062}
Lundberg, S.~M., and Lee, S.-I.
\newblock 2017.
\newblock A unified approach to interpreting model predictions.
\newblock In Guyon, I.; Luxburg, U.~V.; Bengio, S.; Wallach, H.; Fergus, R.;
  Vishwanathan, S.; and Garnett, R., eds., {\em Advances in Neural Information
  Processing Systems 30}. Curran Associates, Inc.
\newblock  4765--4774.

\bibitem[\protect\citeauthoryear{Mostafazadeh \bgroup et al\mbox.\egroup
  }{2016}]{mostafazadeh2016corpus}
Mostafazadeh, N.; Chambers, N.; He, X.; Parikh, D.; Batra, D.; Vanderwende, L.;
  Kohli, P.; and Allen, J.
\newblock 2016.
\newblock A corpus and cloze evaluation for deeper understanding of commonsense
  stories.
\newblock In {\em Proceedings of the 2016 Conference of the North American
  Chapter of the Association for Computational Linguistics: Human Language
  Technologies},  839--849.

\bibitem[\protect\citeauthoryear{Ribeiro, Singh, and Guestrin}{2016}]{lime}
Ribeiro, M.~T.; Singh, S.; and Guestrin, C.
\newblock 2016.
\newblock "why should {I} trust you?": Explaining the predictions of any
  classifier.
\newblock In {\em Proceedings of the 22nd {ACM} {SIGKDD} International
  Conference on Knowledge Discovery and Data Mining, San Francisco, CA, USA,
  August 13-17, 2016},  1135--1144.

\bibitem[\protect\citeauthoryear{Robinson}{1965}]{robinson1965machine}
Robinson, J.~A.
\newblock 1965.
\newblock A machine-oriented logic based on the resolution principle.
\newblock {\em Journal of the ACM (JACM)} 12(1):23--41.

\bibitem[\protect\citeauthoryear{Rockt{\"a}schel and
  Riedel}{2017}]{rocktaschel2017end}
Rockt{\"a}schel, T., and Riedel, S.
\newblock 2017.
\newblock End-to-end differentiable proving.
\newblock {\em arXiv preprint arXiv:1705.11040}.

\bibitem[\protect\citeauthoryear{Rudinger \bgroup et al\mbox.\egroup
  }{2020}]{rudinger2020thinking}
Rudinger, R.; Shwartz, V.; Hwang, J.~D.; Bhagavatula, C.; Forbes, M.; Le~Bras,
  R.; Smith, N.~A.; and Choi, Y.
\newblock 2020.
\newblock Thinking like a skeptic: Defeasible inference in natural language.
\newblock In {\em Findings of the Association for Computational Linguistics:
  EMNLP 2020},  4661--4675.

\bibitem[\protect\citeauthoryear{Sharma \bgroup et al\mbox.\egroup
  }{2015a}]{sharma2015identifying}
Sharma, A.; Vo, N.; Aditya, S.; and Baral, C.
\newblock 2015a.
\newblock Identifying various kinds of event mentions in k-parser output.
\newblock In {\em Proceedings of the The 3rd Workshop on EVENTS: Definition,
  Detection, Coreference, and Representation},  82--88.

\bibitem[\protect\citeauthoryear{Sharma \bgroup et al\mbox.\egroup
  }{2015b}]{sharma2015towards}
Sharma, A.; Vo, N.~H.; Aditya, S.; and Baral, C.
\newblock 2015b.
\newblock Towards addressing the winograd schema challenge—building and using
  a semantic parser and a knowledge hunting module.
\newblock In {\em Twenty-Fourth International Joint Conference on Artificial
  Intelligence}.

\bibitem[\protect\citeauthoryear{Vaswani \bgroup et al\mbox.\egroup
  }{2017}]{vaswani2017attention}
Vaswani, A.; Shazeer, N.; Parmar, N.; Uszkoreit, J.; Jones, L.; Gomez, A.~N.;
  Kaiser, {\L}.; and Polosukhin, I.
\newblock 2017.
\newblock Attention is all you need.
\newblock In {\em Advances in neural information processing systems},
  5998--6008.

\bibitem[\protect\citeauthoryear{Wolf \bgroup et al\mbox.\egroup
  }{2020}]{wolf2020transformers}
Wolf, T.; Debut, L.; Sanh, V.; Chaumond, J.; Delangue, C.; Moi, A.; Cistac, P.;
  Rault, T.; Louf, R.; Funtowicz, M.; et~al.
\newblock 2020.
\newblock Transformers: State-of-the-art natural language processing.
\newblock In {\em Proceedings of the 2020 conference on empirical methods in
  natural language processing: system demonstrations},  38--45.

\end{thebibliography}

\end{document}